# Bayesian Meta-Reasoning: Determining Model Adequacy from Within a Small World

Kathryn Blackmond Laskey
Department of Systems Engineering and C$^3$I Center
George Mason University
Fairfax, VA 22030
klaskey@gmu.edu

## Abstract

This paper presents a Bayesian framework for assessing the adequacy of a model without the necessity of explicitly enumerating a specific alternate model. A test statistic is developed for tracking the performance of the model across repeated problem instances. Asymptotic methods are used to derive an approximate distribution for the test statistic. When the model is rejected, the individual components of the test statistic can be used to guide search for an alternate model.

## 1 INTRODUCTION

Bayesian methods have been gaining in popularity among researchers across a broad variety of disciplines concerned with the problem of reasoning under uncertainty. Bayesian theory can account for many important aspects of human scientific reasoning, including Occam's razor and the ability to make causal inferences without explicit randomization (e.g., Howsen and Urbach, 1989; Jeffreys and Berger, 1992). Bayesian theory gives a satisfying account of the process by which evidence is used to revise one's degrees of belief in a set of explicitly enumerated hypotheses, given a joint probability model on evidence and hypotheses. Hierarchical Bayesian models permit the use of data to revise beliefs about which of class of models gave rise to the data. But there are two essential aspects of intelligent reasoning under uncertainty which standard accounts of Bayesian theory fail to handle. While Bayes' Theorem tells us how to compare explicitly enumerated hypotheses in the light of evidence, it gives no guidance on enumerating new hypotheses and provides no mechanism for rejecting an hypothesis without an explicitly enumerated alternative. Both these capabilities are essential to the ability to reason flexibly on high-dimensional problems for which it is impractical to specify a single, static probability model in advance of seeing any data.

This paper focuses on the problem of model failure diagnosis. Several authors have proposed heuristic approaches to model failure diagnosis (e.g., Jensen, et al., 1990; Habbema, 1976). Laskey (1991) suggested that the agent construct artificial "straw models" as foils against which to compare the current model. But no guidance is given on how to select the straw model, and no distributional theory for the resulting test statistic is available. Other authors (Herskovits and Cooper, 1990; Laskey, 1992; Pearl, 1991) suggest sequentially generating a sequence of models and selecting the best according to some goodness criterion. For Laskey, the best model has the highest posterior probability. Herskovits and Cooper select models based on entropy. Pearl assumes the complete probability distribution (at least the complete conditional independence structure) is known and selects the simplest model which satisfies the conditional independence constraints. These approaches all require explicitly enumerating a comparison model before dropping the current one.

## 2 HYPOTHESIS TESTS AND THE BOUNDED BAYESIAN

Classical statistics has developed a number of procedures for model failure diagnosis, including goodness-of-fit test statistics and heuristic methods for detecting outliers. In the classical hypothesis testing paradigm, one postulates a family of statistical models, one of which is assumed to have given rise to the data. Some subfamily of this family of models is singled out as the null hypothesis (the hypothesis to be assumed unless the data discredit it). A test statistic is developed whose distribution is known, at least approximately, under the null hypothesis. Prior to observing the data, one specifies a range within which it is highly probable that this test statistic will fall. If the observed value of the test statistic falls outside the range, the null hypothesis is rejected. The *level* of the test is the probability that one will reject a true null hypothesis (a Type I error). The *power* of the test against a given alternative hypothesis is the probability that one will reject the null hypothesis if the alternative hypothesis is true (no Type II error).

Bayesian statisticians have criticized classical statistical methods because their use leads to incoherence (violation of the axioms of subjective expected utility theory) and because the conclusions one can draw from a classical hypothesis test are not those needed for decision making. Decision makers generally are interested in how strongly one should believe the null hypothesis relative to the alternative hypothesis. Classical tests do not directly answer this question. A classical hypothesis test reports the *a priori* probability (before seeing the observed value of the test statistic) under the null hypothesis that a value as extreme as the actual observation would occur. An extreme observation is interpreted as casting doubt on the null hypothesis, but no inference about actual degrees of belief in null or alternative hypothesis is warranted.

Bayesian inference methods do answer the decision maker's question -- they report degrees of belief in the null hypothesis conditional on the observed value of the test statistic. Many classical methods can be interpreted as approximate Bayesian methods when the sample is sufficiently large that the data swamp the prior distribution. Thus, Bayesian statisticians often use classical methods, although they regard the classical interpretation of the results as incorrect. But unlike classical methods, Bayesian methods require explicit computation of the likelihood of the data under the alternative models and a probability distribution for the current and alternative models.

The approach I propose below is suggestive of a classical hypothesis test, but my interpretation is neither classical nor traditional Bayesian. The traditional Bayesian has nothing to say about a problem in which one cannot explicitly enumerate and compute data likelihoods and a prior distribution for one's alternate hypothesis. But the classical statistician eschews the subjectivist interpretation of probability, which is the interpretation I take in this paper.

The ultimate goal of the research program of which this paper is a part is a theory of robust approximate Bayesian inference on problems in which explicit enumeration and computation of a fully adequate probability model is impractical or impossible. At any given time, the agent reasons within an approximate "small world" model (which may or may not be explicitly Bayesian). This model is believed at the time it is adopted to be an adequate approximation to a "larger world" model which more fully represents the agent's beliefs. The role of a test statistic is to serve as a computationally simple measure of the adequacy of the approximation. Such a test statistic can serve to alert a system that it needs to devote more resources to explicit enumeration of alternate models, to suggest directions for model search, or to trigger a system to turn to a human operator for assistance.

There is an explicitly Bayesian interpretation of the model failure diagnosis approach I propose. In this view, the agent's current model is viewed as an approximation to a probability distribution over a family of models. This full probability distribution is too complex to be computed explicitly. Given repeated observations, the Central Limit Theorem can be used to derive an approximate distribution for the test statistic for any model in the family. This enables an approximate meta-level Bayesian analysis of the adequacy of the current model.

## 3  A TEST STATISTIC

Let $X_1, ... X_n$ be a sample from a t-dimensional multinomial distribution with parameters 1 and $\pi$. That is, each $X_i$ is a vector of t-1 zeros and a single 1, with the probability of a 1 in the ith position equal to $\pi_i$. The current model for the probability distribution of the $X_i$ is that they are independent and identically distributed with probability equal to a fixed vector p. The goal is to evaluate the adequacy of p as an approximation to $\pi$. I assume that all components of both p and $\pi$ are strictly greater than zero.

This situation arises in many current applications of Bayesian models to expert systems. In these models, $X_i$ indicates which cell in a cross-classified table the ith observation falls into. The current model p is estimated by eliciting from an expert a network encoding independence assumptions among the classifying variables and conditional probability distributions of variables given their neighbors (a Bayesian network). The probability vector p may also be fixed in advance when the probability values are estimated from a sample different from the one used to evaluate the model.

Define the random variables $Y_i$ as follows:

$$Y_i = \sum_{k=1}^{t} X_{ik} \log(p_k) \qquad (1)$$

The variables $Y_i$ are independent and identically distributed with mean

$$E[Y_i] = \mu = \sum_{k=1}^{t} \pi_k \log(p_k). \qquad (2)$$

If $p=\pi$, the mean of $Y_i$ is equal to

$$\mu_* = \sum_{k=1}^{t} \pi_k \log(\pi_k). \qquad (3)$$

The quantity $\mu_\pi$ is called the entropy of the distribution $\pi$. The $Y_i$ are called the *logarithmic scores* of the data $X_i$ under the model p. The logarithmic score is a *strictly proper* scoring rule -- that is, the score is maximized by setting p equal to the correct probability distribution $\pi$. The difference $\mu_\pi - \mu_r$ which is always positive, is often used as a measure of the distance between the probability distributions p and $\pi$.

One could estimate (3) by the entropy $\mu_p$ of the distribution p, which is obtained by replacing $\pi_k$ by $p_k$ in (3). An efficient algorithm for computing $\mu_p$ in Bayesian networks is given by Herskovits and Cooper (1990).

The Central Limit Theorem states that the sample mean of a sample of independent and identically distributed random variables approximates a normal distribution as the sample size becomes large. That is,

$$Z = \sqrt{n}\frac{\overline{Y} - \mu}{\sigma} \Rightarrow \mathcal{N}(0,1) \quad (4)$$

where $\sigma$ is the standard deviation of the $Y_i$ and the symbol $\Rightarrow$ denotes convergence in distribution. Exact computation of the standard deviation $\sigma$ is much less tractable than computation of $\mu$, but its value can be estimated by the sample standard deviation. A classical confidence interval for $\mu$ is given by:

$$\frac{|\overline{Y} - \mu|}{S/\sqrt{n}} \leq z_\alpha \quad (5)$$

where S is the sample standard deviation and $z_\alpha$ is the $\alpha$th percentage point of the normal distribution. The interval (5) can also be interpreted as an approximate posterior credible interval for the unknown value of $\mu$, assuming a prior distribution for $\mu$ that is reasonably flat in the vicinity of $\overline{Y}$.

The model failure test statistic W is defined by

$$W = \frac{|\overline{Y} - \mu_p|}{S/\sqrt{n}} \quad (6)$$

A model failure test based on W calls the model into question when (6) is greater than $z_\alpha$ for some specified $\alpha$ (that is, when $\mu_p$ falls outside the credible interval for $\mu$). For a log-linear model such as a Bayesian network, tests based on W can detect incorrect estimates of the sufficient statistics of the model (the joint probability distribution for each node and its parents) but cannot detect incorrect model structure. That is, when p is estimated by maximum likelihood or an asymptotically equivalent method (as, for example, in Herskovits and Cooper, 1990), the mean (2) converges in probability to the entropy $\mu_p$ whether or not the model is correct.

## 4 DETECTING INCORRECT STRUCTURE

To test for incorrect model structure, I develop a test based on the conditional entropy. Assume that the observations $X_i$ are cross-classified on a set of s variables: $X_i = (Q_{1i}, ...Q_{qi})$, where $Q_{ri}$ denotes the value for the rth classification variable on observation i (the $Q_r$ are the nodes in the Bayesian network). The total number of categories t is then equal to $t_1 \cdots t_s$, where $t_r$ is the number of possible values for classification variable r.

Let $(p|q_r)$ denote the conditional distribution of the $X_i$ given that $Q_r = q_r$. Denote the entropy of this distribution by $\mu_{p|q_r}$. For those i for which $Q_{ri}=q_r$, define

$$Y_{i|q_r} = \sum_k X_{ik} \log((p|q_r)_k) \quad (7)$$

the logarithmic score under the conditional distribution given $Q_{ri} = q_r$. Exactly as above, the sample average $\overline{Y}_{i|q_r}$ is asymptotically normally distributed (where the sample mean and standard deviation are taken over values for which $Y_{i|q_r}$ is defined), and a test statistic $W_{q_r}$ can be constructed in analogy to (6).

This statistic $W_{q_r}$ can be used to test model structure. One would not expect the conditional logarithmic score under $(p|q_r)$ to converge to the conditional entropy $\mu_{p|q_r}$ when there are conditional dependencies involving the node $Q_r$ that are not represented in the distribution $(p|q_r)$.

The conditional entropy can be computed fairly easily. In the Lauritzen and Spiegelhalter algorithm as described in Neapolitan (1990), evidence $Q_r=q_r$ is absorbed into an existing clique tree by creating a new clique tree without the node $Q_r$ and setting clique potentials in the new clique tree to values from the original clique tree corresponding to $Q_r=q_r$. The conditional entropy is just the entropy of this modified clique tree.

A test statistic based on (7) can be computed for each cross-classifying variable $Q_r$. I have derived an asymptotic distribution only for the marginal distribution of each test statistic. The full joint distribution would be difficult to derive. Thus, only variable-by-variable intervals are available, not a credible region within which the vector of expected scores is be expected to fall. It should be noted that when there is a large number of cross-classifying variables, one would expect some of the tests to suggest model failure even when the model is correct. Nevertheless, the pattern of values of the $W_{q_r}$ can serve as a useful heuristic indicator of potential problems with the model.

When model failure diagnosis is to be used to trigger search for alternate models, values of the vector of test statistics can be used to guide search. If the test statistic based on $\bar{Y}_{i|q_r}$ exceeds its threshold, this can serve as an indicator that a new arc may be needed between node $Q_r$ and some other node.

## 5 INCOMPLETE DATA

In many applications, it is important to be able to diagnose model failure from incomplete data. In applications of Bayesian networks to expert systems, one usually observes only some of the variables and the goal is to infer marginal probabilities for the other variables given values for the observed variables. It is important to be able to diagnose model inadequacy from such incomplete data.

As before, let $X_i$ denote the ith observation, but assume that only some of the $Q_{ri}$ are observed. Let $X_i^*$ denote the complete (unobserved) data. Define

$$Y_i^* = \sum_{k=1}^{t} X_{ik}^* \log(p_k) ; \qquad (8)$$
$$Y_i = E[Y_i^* | X_i].$$

The $Y_i$ are independent with mean $\mu$. When the same variables are observed for all cases, the $Y_i$ are also identically distributed, and the theory developed above applies. When different data patterns are observed for different cases, the standard deviation of $Y_i$ differs with i, and distributional theory for W or $W_{p|q_r}$ is not available. Nevertheless, these statistics could still serve as useful heuristic indicators. Because of the additive decomposition of $\log(p_k)$, computation of $Y_i$ can be performed by a minor modification of the Lauritzen and Spiegelhalter algorithm.

## 6 DISCUSSION

A set of test statistics for diagnosing model inadequacy was developed. When a sample of observations is available and data are complete, an approximate posterior credible interval for each test statistic can be derived. When data are incomplete, the expected value of the test statistics can be computed.

Extreme values of the test statistics can be used to initiate search for an alternate model. In addition, the pattern of which statistics exceed threshold can be used to guide search for an alternate model.

### Acknowledgments

This research was partially funded by a grant from the Virginia Center for Innovative Technology to the Center of Excellence in Command, Control, Communications and Intelligence at George Mason University.

## References

Habbema, J.D.F. (1976). Models for Diagnosis and Detection of Combinations of Diseases. In deDombal et al. (eds), *Decision Making and Medical Care*. Nor h Holland.

Herskovits, E. and Cooper, G. (1990) Kutato: An Entropy-Driven System for Construction of Probabilistic Expert Systems from Databases. In *Proceedings of the Sixth Conference on Uncertainty in Artificial Intelligence*, Boston, MA: Association for Uncertainty in Artificial Intelligence, Mountain View, CA.

Howsen, C. and Urbach, P. (1989) *Scientific Reasoning: The Bayesian Approach*. LaSalle, IL: Open Court.

Jensen, F.V., Chamberlain, B., Nordahl, T. and Jensen, F. (1990). Analysis in HUGIN of Data Conflict. In *Proceedings of the Sixth Conference on Uncertainty in Artificial Intelligence*. Mountain View, CA: Association for Uncertainty in Artificial Intelligence, pp. 546-554.

Jeffreys, W.H. and Berger, J.O. (1992) Occam's Razor and Bayesian Analysis. *American Scientist*, Vol. 80, p. 64-72.

Laskey, K. B. (1991) Conflict and Surprise: Heuristics for Model Revision. In D'Ambrosio, B.D., P. Smets and P.P Bonissone (eds.). *Uncertainty in Artificial Intelligence: Proceedings of the Seventh Conference*. San Mateo, CA: Morgan Kaufmann.

Laskey, K.B. (1992) The Bounded Bayesian, this volume.

Neapolitan, R.E. (1990) *Probabilistic Reasoning in Expert Systems: Theory and Algorithms*, New York: Wiley.

Pearl, J. (1991) A Theory of Inferred Causation, In Allen, J.A., Fikes, R. and Sandewall, E. (Eds.) *Principles of Knowledge Representation and Reasoning: Proceedings of the Second International Conference*. San Mateo, CA: Morgan Kaufmann, pp. 441-452.